# Places in the Wild:

# A Large, High-Resolution RAW Photograph Dataset for Ecologically Valid Vision Research

Michelle R. Greene
Barnard College, Columbia University

mgreene@barnard.edu

# Abstract

Large image datasets have accelerated progress in cognitive neuroscience and computer vision. However, most datasets are low-resolution, internet-sourced JPEGs with unknown capture conditions and limited spatial context. Places in the Wild is a dataset of 67,574 high-resolution photographs collected in situ across 810 physical locations spanning 260 basic-level scene categories, including indoor, urban, and natural environments. At each location, a 45-megapixel Canon EOS R5 mounted on a panoramic tripod captured 72 images at 5-degree horizontal intervals plus 12 images at varying elevations, yielding dense 360-degree viewpoint sampling. All images were recorded simultaneously as 14-bit RAW (CR3) files and compressed JPEGs, preserving sensor-level detail for analyses of luminance, contrast, color, and other image statistics. The dataset is accompanied by complete EXIF metadata and a suite of image-quality metrics. Places in the Wild supports research on viewpoint-dependent recognition in humans and models, training and evaluation of scene-understanding systems under realistic conditions, characterization of natural scene statistics, and experiments requiring near-full-field visual displays.

## Background and Summary

The statistics of our visual environment shape perception, behavior, and brain organization, but much of what we know about visual experience rests on datasets that are low-resolution, limited in diversity, or curated from online content. Large-scale image corpora such as ImageNet [1], SUN [2], and Places [3] have been instrumental in advancing our understanding of object and scene recognition. However, these datasets are typically constructed from internet-sourced JPEGs of modest resolution, with uncontrolled compression artifacts, unknown camera settings, and limited information about the spatial context from which each image was captured. To make progress in human and machine vision, we need datasets that better approximate the richness and structure of everyday experience. Natural vision unfolds through embodied movement across time and space, providing the visual system with continuous variation in viewpoints.

To address these limitations, we present *Places in the Wild*, a new dataset of high-resolution, real-world scene photographs designed for research in visual neuroscience, computer vision, and computational modeling. The dataset comprises over 67,000 RAW-format images, each captured with a 45-megapixel Canon EOS R5 full-frame mirrorless camera at 5-degree intervals, providing 360-degree coverage across over 800 unique locations. These locations span 260 basic-level scene categories, including both indoor and outdoor environments such as bedrooms, train stations, forests, and parking garages.

A standardized data collection protocol was employed to ensure spatial consistency across locations. At each site, a professional tripod with a protractor mount was used to rotate the camera along a horizontal arc, acquiring a complete panoramic sweep from a single fixed vantage point. All images were recorded in RAW (CR3) format, preserving sensor-level detail and enabling high-precision analyses of luminance, contrast, color balance, and other image statistics.

The dataset contains over three trillion pixels, making it approximately 11 times larger than ImageNet by total pixel count. To our knowledge, it is the largest publicly available dataset of high-resolution, real-world photographs with dense viewpoint sampling. *Places in the Wild* is thus suitable for a wide range of use cases, including analyzing viewpoint-dependent recognition performance in humans and models, training and evaluating scene understanding systems under real-world conditions, measuring low-level visual statistics across diverse environments, and experimental applications that require near-full-field displays.

This paper describes the data acquisition process, image formatting and storage conventions, metadata schema, and validation procedures. All materials are made available in accordance with open data best practices, and the dataset is released under a permissive license (CC-BY-NC-SA) to facilitate reuse by researchers across disciplines.

# Methods

**Hardware:** Photographs were taken with a Canon EOS-R5 mirrorless full-frame camera equipped with a Canon RF 24-105mm F4 L IS USM lens. We mounted the camera on an Oben AT-3565 aluminum tripod featuring a built-in panoramic base with 5-degree protractor markings, enabling precise angular rotation at fixed increments. At each capture point, the lens was fixed to a 24 mm focal length (unless otherwise specified). For each image, the camera simultaneously recorded a 45-megapixel RAW file (.CR3, 8192 × 5464 pixels, 14-bit depth) and a compressed JPEG version. Exposure settings—including ISO, aperture, shutter speed, and white balance—were automatically configured using Canon's Scene Intelligent Auto mode. See the Technical Validation section for further analysis of EXIF metadata.

**Protocol:** At each data collection site, the tripod was set to a height of approximately 128 cm, and the camera was leveled using the tripod's spirit level. A total of 72 photographs were captured in a single horizontal rotation, with one image taken every 5 degrees. After completing the horizontal sweep, the camera was removed from the tripod, and an additional 12 photographs were taken at varying elevation angles, beginning with a downward-facing image and ending with a view pointing directly upward toward the ceiling or sky. See Figure 1 for a set of example images. As a result, most locations contain 84 images in total. However, a small subset may contain fewer images due to safety constraints (e.g., obstructed walkways or unstable footing) or camera automation failure in visually uniform environments, such as blank white walls or featureless surfaces, where Scene Intelligent Auto was unable to lock exposure. A total of 67,574 images were collected across 810 different locations.

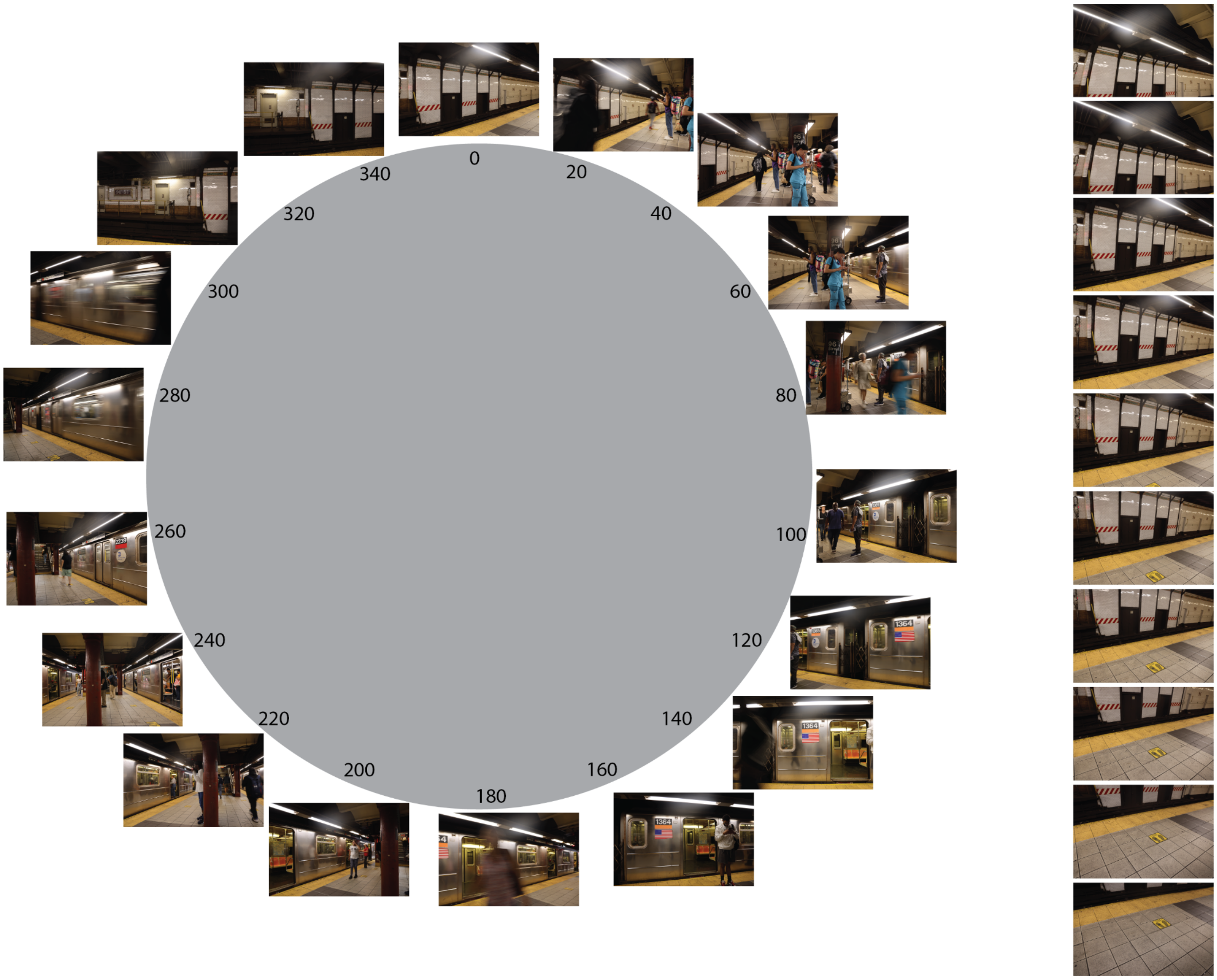

**Figure 1: (Left) A selection of viewpoints sampled at one location (subway platform). Although viewpoints were captured at 5° increments, we are displaying 20° increments for clarity. (Right) A selection of the 12 viewpoints that varied in elevation from the same location.**

**Scene Taxonomy:** Our goal was to sample a broad and representative set of real-world locations. We started with the set of 365 categories from the Places dataset [3], which are derived from basic-level concepts frequently used in natural language and human labeling tasks. From this initial set, we excluded 10 categories that posed safety hazards for human observers or camera equipment (e.g., *volcano*), 20 categories that raised privacy or security concerns (e.g., *operating room*), and 35 categories that could not be practically accessed from the greater New York City area (e.g., *glacier*). Additionally, we merged categories that were highly semantically similar (e.g., “bedroom” and “bedchamber”) to reduce overlap and ensure each label reflected a distinct real-world concept. This yielded a preliminary set of 241 categories. To expand coverage, we incorporated 20 additional scene categories from the SUN dataset [2] and 10 categories suggested by lab members, bringing the total to 271. Ultimately, we were unable to obtain suitable locations for 11 of these categories, yielding a final set of 260 basic-level scene categories (see the Supplementary Materials for a complete list). Figure 2 shows example categories across indoor, urban, and natural environments.

**Sampling Protocol:** We aimed to sample at least three independent physical locations per category whenever feasible. Due to variations in availability and other logistical constraints, the number of locations per category varied, with the dataset averaging 3.11 locations per category (median = 3, range = 1–5). Each location was treated as a distinct capture site, with no reuse of the same physical space across categories. Locations were selected to maximize scene diversity within each category, accounting for architectural variation, lighting conditions, and occupancy status (e.g., occupied vs. vacant offices, public vs. residential stairways).

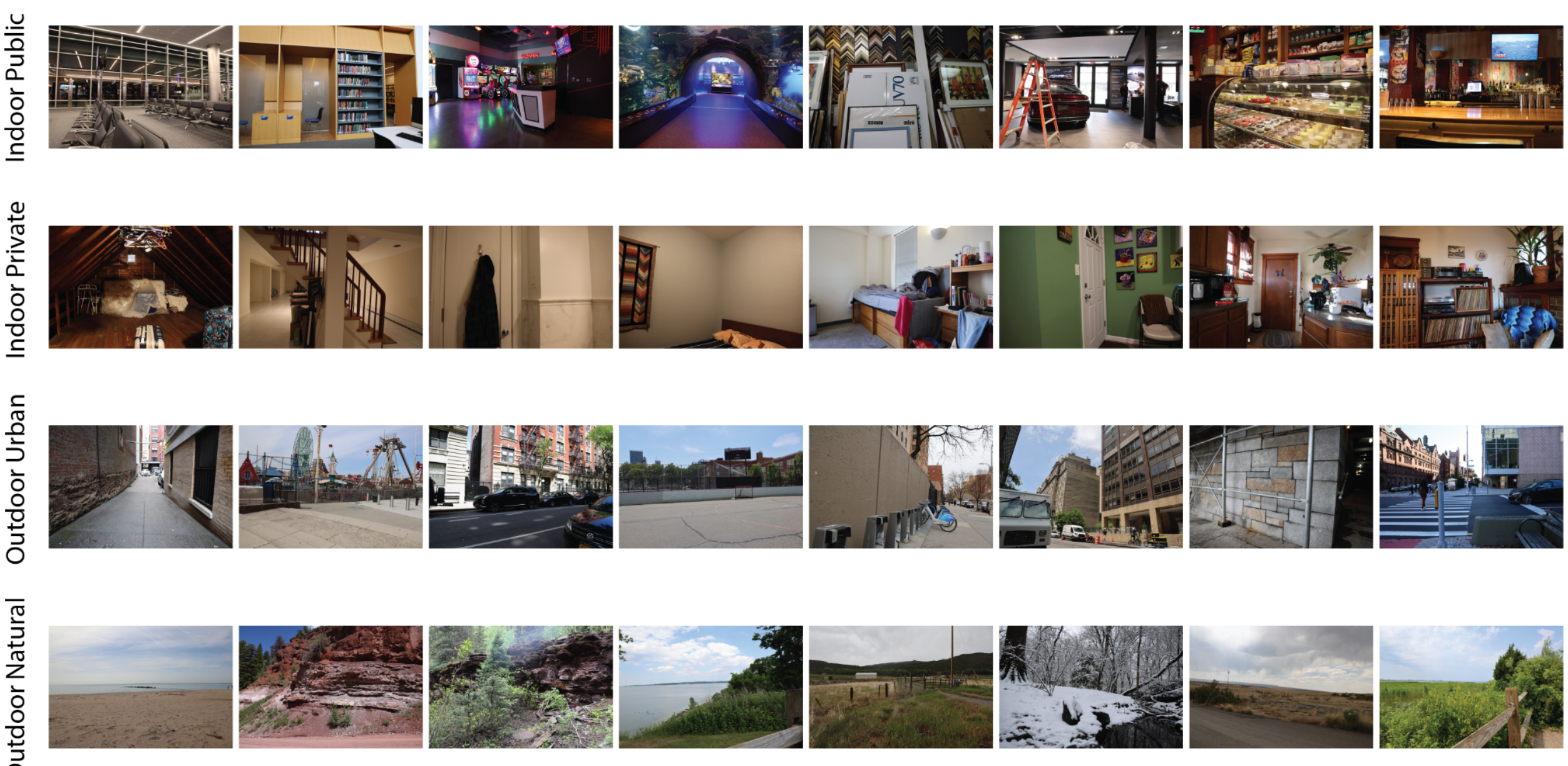

**Figure 2: Sample images from a variety of locations. The top two rows depict indoor locations, and the bottom two rows depict outdoor locations.**

**Ethics Statement:** Photography focused on physical environments rather than individuals. All private residences and businesses were photographed with the explicit verbal permission of the owner or manager, who consented to the public release and redistribution of the resulting images. Photographs were framed to avoid people wherever possible. No personally identifying metadata (owner names, exact addresses, or GPS coordinates) is included in the released files.

# Data Records

Places in the Wild comprises 67,574 images from 810 locations, distributed across 260 basic-level scene categories. Of these, 394 locations (48%) are indoor, 235 (29%) are urban, and 181 (22%) are natural.

The dataset can be accessed as a torrent file, accessible at this URL: https://academictorrents.com/details/a1d810fbb54c21dfe8a68538d917c668e48663de. The torrent contains a single .tar file of the entire dataset (2.62 TB total). Inside the uncompressed folder, images are organized by basic-level category, and within each category folder, the separate locations are in subfolders. Within each subfolder, images are named sequentially. For example, canyon_loc3_0.CR3 is the first image in the third canyon location. All files are in CR3 format (Canon's RAW format) and JPG.

# Technical Validation

**Viewpoint Diversity**

Some scene categories, such as *forest*, are evident from nearly all viewpoints. Others are characterized by regions visible only from a subset of locations (e.g., an outdoor view of a *library*). To quantify the semantic variability of each basic-level scene category across viewpoints, we computed a category entropy metric for each location. We used OpenAI's CLIP model (ViT-B/32 [4]) to assign basic-level scene labels to each image using zero-shot classification. Specifically, for each image, we computed the cosine similarity between the image's embedding and each category's text embedding, using the list of 260 basic-level scene categories from the dataset's taxonomy. The category with the highest similarity was assigned as the predicted label for that image. At each location, we computed the Shannon entropy of predicted category labels across all viewpoints. This metric reflects the degree of semantic coherence across different angles from the same physical location. Lower entropy values indicate locations with self-similar semantics across all viewpoints, while higher values indicate greater semantic variability. Finally, we averaged entropy across all locations for each basic-level category.

Figure 3 shows examples of high- and low- entropy locations. Overall, we found that category entropy ranged from 0.15 bits for fabric stores to 4.07 bits for art schools. Generally, outdoor scenes had greater category entropy than indoor (2.36 bits versus 2.15 bits, $t(237.6) = -2.26$, $p=0.025$), and urban locations had higher category entropy than non-urban (2.5 bits versus 2.15 bits, t(199.9) = 4.04, p=7.5e-5). However, natural landscapes did not differ significantly from other categories in category entropy (2.15 bits versus 2.29 bits, t(94.8) = -1.32, p = 0.19).

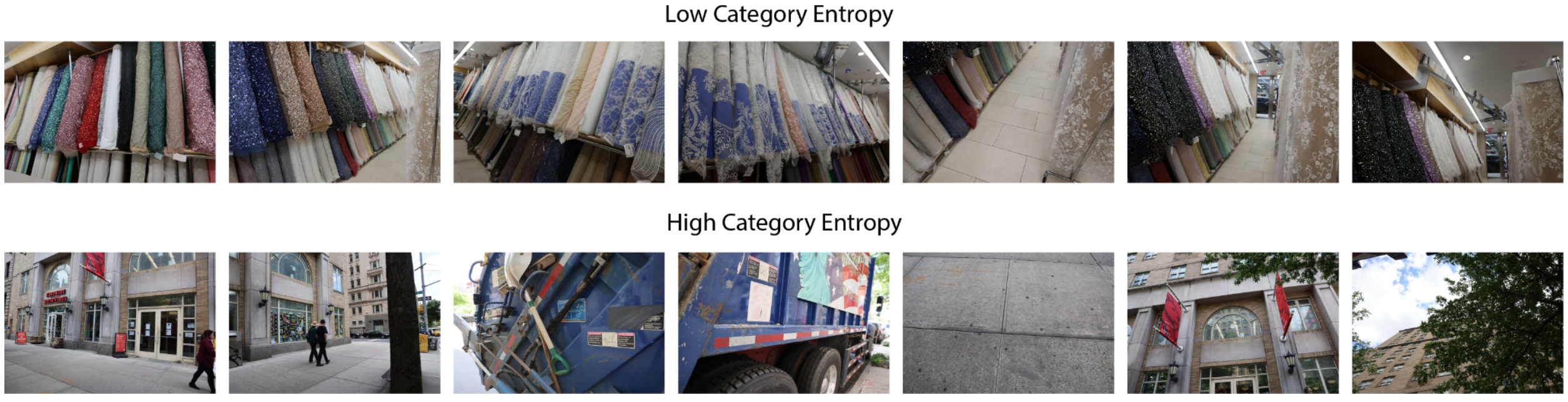

**Figure 3: (Top) Example of viewpoints from a low entropy location. Each viewpoint was labeled as a "fabric store". (Bottom) Example viewpoints from a high entropy location.**

**Visualizing Dataset Diversity**

To quantify the visual diversity within our dataset, we extracted high-level visual features using a pretrained ResNet-50 convolutional neural network [5] from the PyTorch library, trained on ImageNet [1]. Specifically, we used the final average pooling layer (2048-dimensional vector). A forward hook was registered to this layer to capture features during inference.

We preprocessed images according to the standard ResNet pipeline: resizing to 256 pixels on the shorter edge, center-cropping to 224 x 224, and normalizing to the ImageNet mean and standard deviation. To minimize memory overhead, features were extracted in batches with gradients disabled. We reduced the dimensionality of the resulting feature matrix using principal component analysis (PCA), retaining the first 10 components. We visualized the space using t-SNE (see Figure 4A).

To estimate the semantic diversity of PitW, we first obtained basic-level categories for each image using GPT-4o (OpenAI). We then encoded each description using the MPNet embedding [6]. As with the visual diversity analysis, we reduced the dimensionality of the embeddings using PCA, then computed t-SNE to visualize the space. Figure 4A shows a t-SNE representation of the visual features from Resnet-50, and 4B shows a similar plot of semantic features from MPNet. We found that both visual and semantic features separated the indoor from outdoor images, and that visual features clearly separated natural outdoor spaces from urban ones, while weaker separation existed between these categories for the semantic features. Neither visual nor semantic features separated the private from public indoor spaces.

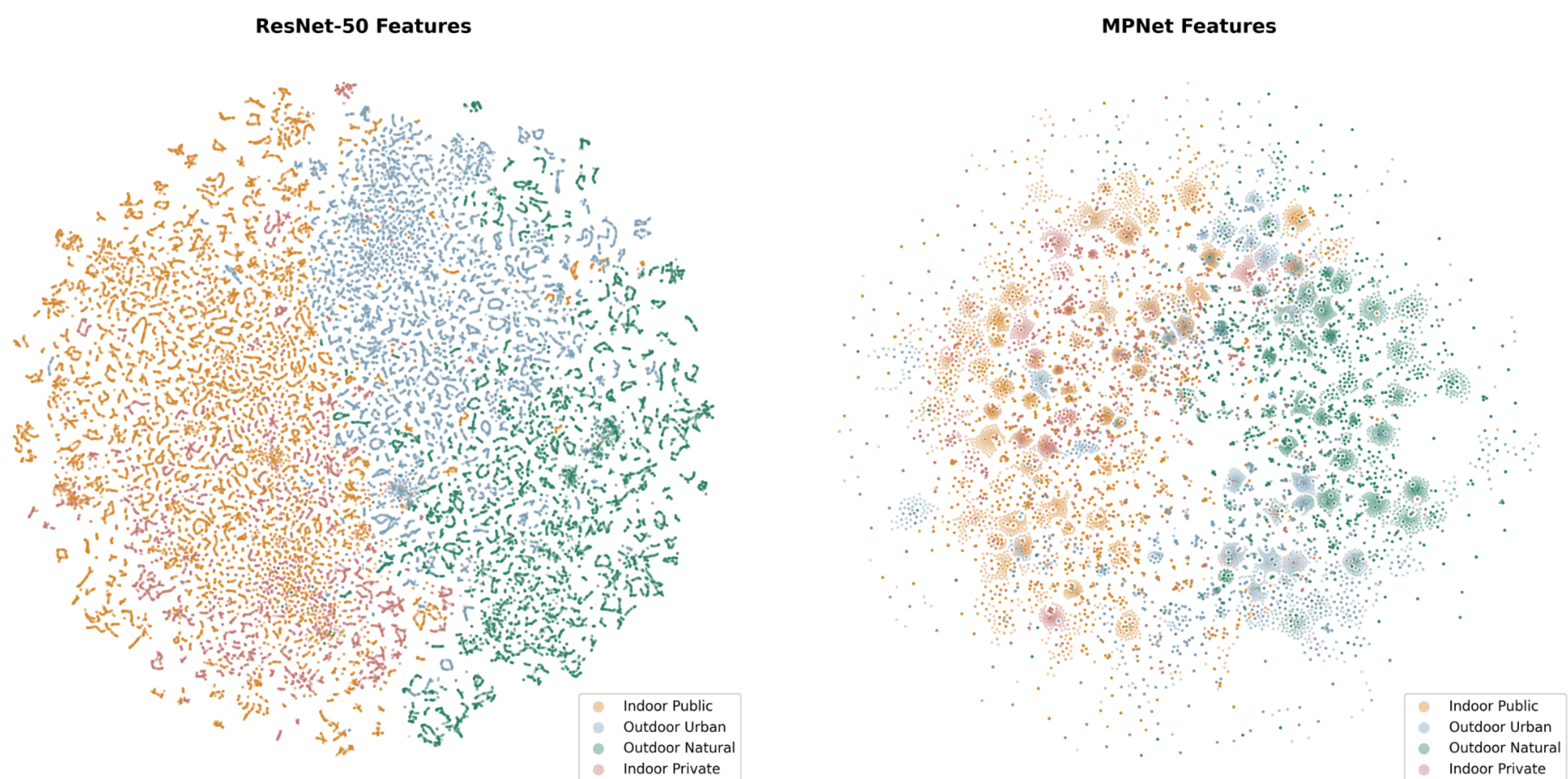

**Figure 4: (Left) t-stochastic neighbor embedding (t-SNE) embedding for visual features from the Resnet-50 convolutional neural network. Images are color-coded according to superordinate level categories. (Right) t-SNE embedding for semantic features from the MPNet sentence embeddings, color-coded by superordinate level categories.**

### Dataset Quality and Metadata

We analyzed EXIF metadata for all photographs in the dataset, including aperture (FNumber), shutter duration (exposure time), sensor sensitivity (ISO), focal length, and exposure program mode (see Table 1). Nearly all photographs were captured using Canon's Scene Intelligent Auto mode, with a fixed focal length of 24 mm in 95% of images and a dominant aperture setting of f/4.0, used in 37% of cases. This wide-aperture setting reflects the lens's maximum aperture, minimizing depth-of-field variation and reducing optical sharpness falloff across the dataset. In bright outdoor environments, the camera automatically adjusted exposure by selecting smaller apertures (e.g., f/8 or higher). The distribution of ISO values indicates a bias toward well-lit conditions, with 73% falling between 100 and 400, while 90% of shutter speeds were faster than 1/60 second, minimizing motion blur.

| Feature | Description | Mode | Median | Mean | Range |
|---|---|---|---|---|---|
| FNumber | Aperture (f-stop) | f/4.0 | f/4.5 | f/5.72 | 4.0-16.0 |
| Exposure time | Shutter duration | 0.0167 | 0.0125 | 0.022 | 0.0005-3.20 |
| ISO | Sensor sensitivity | 100 | 200 | 840 | 100-12800 |
| Focal length | Lens focal length (mm) | 24 | 24 | 24.53 | 24-105 |
| Exposure program | Camera mode | Scene intelligent auto (99.3%) | NA | NA | NA |
| Exposure compensation | Manual EV override | 0 | 0 | -0.106 | -0.66-0 |

**Table 1: Summary of camera settings extracted from EXIF metadata. Values reflect aggregated settings across the entire dataset. Most images were taken in Canon's *Scene Intelligent Auto* mode, with a dominant focal length of 24 mm and a modal aperture of f/4.0. ISO values and shutter speeds indicate a strong bias toward well-lit, outdoor environments.**

Next, we used a variety of image-processing metrics to evaluate the perceptual quality of images in Places in the Wild. To quantify image sharpness, we computed the variance of the Laplacian of the grayscale image. The Laplacian operator highlights regions of rapid intensity change (edges), and prior work has shown that higher variance in the Laplacian response correlates with a sharper image [7].

To quantify image noise levels, we applied the fast noise variance estimator proposed by [8]. Each image was first converted to 8-bit grayscale, and a fixed high-pass filter kernel was applied to highlight high-frequency content. This kernel suppresses low-frequency structure while amplifying random pixel-level fluctuations associated with noise. We then computed the mean of the absolute filter response across the image and scaled it by a constant factor $\sqrt{(\pi/2)}/6$, which accounts for the conversion from mean absolute deviation (MAD) to standard deviation and the kernel's normalization factor. The result is a per-image estimate of the noise standard deviation (σ), reported on the original 0–255 intensity scale.

We computed the overall colorfulness of the images using the method suggested by [9]. This method computes colorfulness based on the distribution and magnitude of color-channel differences in RGB space. Specifically, we first converted each image to 32-bit RGB and computed the red-green (R–G) and yellow-blue (0.5 × (R+G) – B) opponent color channels. The mean and standard deviation of these channels were then combined into a scalar colorfulness index:

$$C = \sqrt{\sigma_{rg}^2 + \sigma_{yb}^2} + 0.3\ x\ \sqrt{\mu_{rg}^2 + \mu_{yb}^2}$$

where σ and μ denote the standard deviation and mean, respectively, of the red-green (rg) and yellow-blue (yb) differences. This metric reflects both the intensity of chromatic variation and the degree of color deviation from gray, providing a metric of image colorfulness that correlates with human assessments.

Additional luminance-based quality metrics included the mean and median gray levels, the fraction of clipped pixels (i.e., values outside the range of 0 or 255), the robust dynamic range (1st to 99th percentile), and the root mean square (RMS) contrast of the grayscale image.

Finally, we evaluated the perceptual and technical quality of our images using a suite of no-reference image quality assessment (NR-IQA) metrics. We included three metrics, each grounded in slightly different assumptions about the statistics of natural scenes and the nature of perceptual distortions. First, we computed BRISQUE (Blind/Referenceless Image Spatial Quality Evaluator, [10]), which models image quality based on natural scene statistics. For this analysis, we used the OpenCV implementation, which was trained on the LIVE IQA database [11]. Next, we computed the Natural Image Quality Evaluator (NIQE, [12]), a model that estimates image quality without training on human evaluations. It does so by comparing statistical features of the test image to those from a corpus of high-quality natural images. Finally, we included Perception-based Image Quality Evaluator (PIQE, [13]), which estimates perceptual distortion by analyzing block-level sharpness, contrast, and saturation artifacts. NIQE and PIQE were implemented in the Python scikit-image library.

Summary statistics for image quality are found in Table 2. Where available, we provide published interpretive benchmarks in the Comments column to contextualize these values.

| Metric | Median | Min | Max | IQR | Comments |
|---|---|---|---|---|---|
| BRISQUE | 24 | 0 | 141.9 | 14.31 | <45 considered good quality |
| NIQE | 3.6 | 1.68 | 16.1 | 1.13 | < 6 considered good quality |
| PIQE | 45.8 | 4.36 | 85.4 | 15.41 | <20 excellent; 21-35 good; 36-50 fair |
| Laplacian variance (sharpness) | 870.2 | 3.6 | 9873 | 1554 | |
| Noise sigma | 0.008 | 0.0008 | 0.07 | 0.014 | 0-1 scale |
| SNR | 44.4 | 4.95 | 477.7 | 62.4 | |
| Mean luminance | 0.41 | 0.07 | 0.82 | 0.11 | 0-1 scale |
| RMS contrast | 0.23 | 0.03 | 0.44 | 0.06 | |
| Clip fraction | 0.0004 | 0 | 0.43 | 0.005 | <0.02 considered acceptable in professional photography |
| Robust range | 0.88 | 0.14 | 1 | 0.14 | |
| Colorfulness | 27.7 | 2.42 | 98.9 | 18.3 | 15: somewhat colorful; 33: moderately colorful; >45: colorful; >70: very colorful |

**Table 2: Summary of image quality metrics for all 67,574 images in Places in the Wild. For metrics with published interpretive scales, benchmarks are provided in the Comments column.**

Across the no-reference quality metrics, PitW images were generally rated as good to high quality. The median BRISQUE score, 24, and 94% of images were below the threshold of 45 that has been used in applied settings to distinguish acceptable from poor-quality images [14], and is consistent with scores reported for pristine, undistorted natural photographs in the LIVE IQA database [10]. Similarly, the median NIQE score of 3.6 indicates close conformity to the natural scene statistics of high-quality photographic images [12].

### Comparison to other image datasets

In this section, we compare Places in the Wild with several leading scene datasets (ADE-20k [15], LabelMe [16], SUN [17], Places (standard and high-resolution) [3], ImageNet [1], and COCO [18]) in terms of size, visual and semantic diversity, and image quality. Additionally, we compare it to popular large-scale neuroimaging datasets, including NSD [19], THINGS [20], and BOLD-5000 [21].

Table 3 compares PiTW against these leading image datasets. Although PiTW contains fewer images overall than computer vision datasets such as ImageNet and Places, it has the highest median resolution, the only color depth exceeding 8 bits, and the only release in RAW format. PiTW is also the only dataset not scraped from the internet, lending it greater ecological validity.

| Dataset | N Images (thousands) | N Categories | N Pixels (trillions) | Median Resolution | Bit Depth | Format | Sampling | Diversity |
|---|---|---|---|---|---|---|---|---|
| Places in the Wild | 67.6 | 260 | 3.02 | 8192x 5464 | 14 | .CR3, .JPG | In situ | 57.28 |
| SUN | 16.9 | 918 | 0.01 | 425x338 | 8 | .JPG | Internet | 47.04 |
| Places | 1803.4 | 365 | 0.12 | 256x256 | 8 | .JPG | Internet | 53.46 |
| Places-HR | 1803.4 | 365 | 0.66 | 683x512 | 8 | .JPG | Internet | 40.62 |
| ADE-20k | 25.6 | 1105 | 0.03 | 690x534 | 8 | .JPG | Internet | 44.65 |
| LabelMe | 208 | N/A | 0.13 | 500x400 | 8 | .JPG | Internet | 52.75 |
| COCO | 82.8 | 80 | 0.02 | 640x480 | 8 | .JPG | Internet | 34.33 |
| ImageNet | 1281.2 | 1000 | 0.28 | 500x375 | 8 | .JPG | Internet | 46.33 |
| NSD | 73 | 80 | 0.01 | 425x425 | 8 | .JPG | Internet | 33.98 |
| THINGS | 26.1 | 1854 | 0.04 | 800x800 | 8 | .JPG | Internet | 23.55 |
| BOLD-5k | 4.9 | 1288 | 0.002 | 600x500 | 8 | .JPG | Internet | 23.88 |

**Table 3: Comparison of Places in the Wild with leading datasets in human and machine vision.**

To assess the visual diversity of PiTW relative to the other datasets, we conducted the same visual diversity analysis using ResNet-50 features for each dataset. To ensure comparability across datasets of different sizes, we randomly sampled 5000 images from each dataset. We computed the trace of the covariance matrix of the PCA-reduced feature matrix to represent the total feature visual variance of the sample. Places in the Wild exhibited the highest total variance ($\sigma^2 = 57.28$), followed by Places ($\sigma^2 = 53.46$), then LabelMe ($\sigma^2 = 52.75$). COCO had the lowest total variance ($\sigma^2 = 34.33$) among non-neuroimaging datasets, likely due to its object-centric composition and contextual homogeneity. Furthermore, the visual diversity in the three neuroimaging datasets was lower than that of other datasets. These results suggest that PitW spans a broader range of high-level visual content than existing datasets, consistent with its goal of capturing diverse real-world scenes from an egocentric perspective.

## Usage Notes

**Accessing the dataset.** Because of its size (2.62 TB), Places in the Wild is distributed via BitTorrent through Academic Torrents rather than direct download. BitTorrent is a peer-to-peer protocol in which the file is transferred in pieces from multiple sources simultaneously, reducing the load on any single server. Downloading requires a BitTorrent client; we recommend qBittorrent (https://www.qbittorrent.org), which is free, open-source, and available for Linux, macOS, and Windows.

To retrieve the data, install a client, then open the dataset's page at https://academictorrents.com/details/a1d810fbb54c21dfe8a68538d917c668e48663de, download the .torrent file, and open it in the client. When prompted, select a destination drive with at least 3.3 TB of free space. The client will then download the data automatically; the transfer time depends on network speed and the number of active peers, and the client must remain running until the transfer completes. After your download finishes, we encourage you to leave the client running to "seed" the data, which improves dataset availability to all.

**Working with CR3 files.** All images are stored in Canon's RAW format (.CR3), which preserves the unprocessed 14-bit sensor data rather than a demosaiced, gamma-corrected RGB image. This format retains maximum dynamic range and avoids the compression artifacts of JPEG, but not all image viewers can open these files. We recommend opening CR3 files in Python using rawpy (https://pypi.org/project/rawpy/). Code examples for using this tool are provided in the Description section of the Zenodo link. EXIF metadata (Methods, Table 1) can be read directly from the CR3 files with ExifTool (https://exiftool.org) or, in Python, with the exifread or PyExifTool packages.

## Data Availability

The dataset described in this Dataset Descriptor can be accessed as a torrent file, accessible at this URL: https://academictorrents.com/details/a1d810fbb54c21dfe8a68538d917c668e48663de. The citable set with DOI is available via Zenodo [22].

## Code Availability

Scripts that produced the analyses in this Dataset Descriptor can be found on OSF (https://osf.io/djptq/).

## Contributions

MRG conceived of the project, collected data and supervised data collection, analyzed the data, curated the data, acquired funding, and wrote the manuscript.

## Competing Interests

None

## Acknowledgements

Thanks to Hooriya Aamir, Maria Adkins, Kaiki Chiu, Saniya Gaitonde, Vivian Gao, Emily Lo, Amy Nguyen, and Skylar Stadhard for help in taking the photographs. Thanks also to the many families and business owners who allowed us to photograph their spaces.

## Funding

Supported by CAREER 2240815 to MRG.

## Supplementary Materials

List of scene categories: airport terminal, alcove, alley, amphitheater, amusement arcade, amusement park, apartment building, aquarium, archive, arena, art gallery, art school, art studio, athletic field, athletic studio, atrium, attic, auditorium, auto showroom, bakery, balcony (exterior), balcony (interior), ballroom, bar, barber shop, barn, baseball field, basement, basketball court, bathroom, beach, beauty salon, bedroom, beer garden, bicycle racks, biology laboratory, boardwalk, boat deck, bookstore, botanical garden, bowling alley, boxing ring, brewery, bridge, building facade, bus interior, bus depot, bus stop, butchers shop, butte, cabin, campsite, campus, canal, candy store, canyon, car interior, carousel, castle, cathedral, cemetery, chapel, checkout counter, chemistry lab, church (indoor), church (outdoor), classroom, cliff, closet, clothing store, coast, coffee shop, computer room, conference center, construction site, convenience store, corral, corridor, cottage, courthouse, courtyard, creek, crosswalk, delicatessen, department store, desert road, diner (outdoor), dining hall, dog park, dorm room, downtown, driveway, drugstore/pharmacy, dry cleaner, elevator (door), elevator lobby, entrance hall, escalator, estuary, exam room (medical), fabric store, farm, fastfood restaurant, field (cultivated), field (wild), fire escape, fire station, flea market, florist, food court, food truck, forest (broadleaf), forest path, forest road, formal garden, fountain, garage (outdoor), gas station, gift shop, golf course, greenhouse (indoor), greenhouse (outdoor), gymnasium, harbor, hardware store, hayfield, highway, home office, home theater, hotel (outdoor), hotel room, house, ice cream parlor, ice skating rink (indoor), industrial area, inn, Japanese garden, jewelry shop, junkyard, kennel, kindergarten classroom, kitchen, lagoon, lake, laundromat, lawn, lecture room, library (indoor), library (outdoor), lighthouse, liquor store, living room, loading deck, lobby, locker room, lounge, mailroom, market (indoor), market (outdoor), martial arts gym, mezzanine, mosque, motel, mountain, mountain path, mountain snowy, movie theater (indoor), museum (indoor), museum (outdoor), music store, music studio, natural history museum, newsstand, nursery (children), nursing home, ocean, office, office building, office cubicles, orchard, pantry, park, parking garage, parking lot, pasture, patio, pet shop, physics lab, picnic area, pier, pizzeria, playground, plaza, pond, porch, post office, print shop, promenade, pub (indoor), racecourse, railroad tracks, recreation room, repair shop (cars), repair shop (objects), residential neighborhood, restaurant, restaurant patio, river, roof garden, ruin, sandbox, school exterior, science museum, server room, shed, shoe shop, shopfront, shopping mall, shower, ski resort, skyscraper, soccer field, spa, stadium (soccer), stage (indoor), stage (outdoor), staircase, storage room, street, strip mall, subway platform, supermarket, sushi bar, swimming pool (indoor), swimming pool (outdoor), synagogue (outdoor), taxi stand, television room, tennis court, theater backstage, thriftshop, ticket booth, topiary garden, toyshop, train interior, train station, tunnel, utility room, valley, vegetable garden, vending machine, veterinarian office, volleyball court, waiting room, waterfall, wave, yard, youth hostel, zoo.